\documentclass{article}

\usepackage{arxiv}

\usepackage[utf8]{inputenc} 
\usepackage[T1]{fontenc}    
\usepackage{hyperref}       
\usepackage{url}            
\usepackage{booktabs}       
\usepackage{amsfonts}       
\usepackage{nicefrac}       
\usepackage{microtype}      
\usepackage{lipsum}		
\usepackage{graphicx}
\usepackage{doi}

\usepackage{amssymb}
\usepackage{amsmath}
\usepackage{booktabs}
\usepackage{graphicx}

\title{Zero- and Few-shot Named Entity Recognition and Text Expansion in Medication Prescriptions using ChatGPT}

\author{Natthanaphop Isaradech\textsuperscript{1} \\
	\textsuperscript{1}Department of Community Medicine, Faculty of Medicine\\
        Chiang Mai University\\
	\texttt{natthanaphop.isa@cmu.ac.th}
	\And
	Andrea Riedel\textsuperscript{2,3} \\
	\textsuperscript{2}Friedrich-Alexander-Universität Erlangen-Nürnberg, Erlangen\\
        \textsuperscript{3}Medical Center for Information and Communication Technology\\ Universitätsklinikum Erlangen
	\texttt{andrea.riedel@uk-erlangen.de}
        \AND
	Wachiranun Sirikul\textsuperscript{1,4} \\
        \textsuperscript{1}Department of Community Medicine, Faculty of Medicine \\
        Chiang Mai University \\
        \textsuperscript{4}Center of Data Analytics and Knowledge Synthesis for Health Care \\
        \texttt{wachiranun.sir@cmu.ac.th}
	\And
	Markus Kreuzthaler\textsuperscript{5} \\
	\textsuperscript{5}Institute for Medical Informatics\\               Statistics and Documentation\\
        Medical University of Graz \\
        \texttt{markus.kreuzthaler@medunigraz.at}
	\And
	Stefan Schulz\textsuperscript{5} \\
	\textsuperscript{5}Institute for Medical Informatics\\               Statistics and Documentation\\
        Medical University of Graz \\
        \texttt{stefan.schulz@medunigraz.at}
}

\renewcommand{\shorttitle}{\textit{Pre-print submitted to arXiv}}

\hypersetup{
pdftitle={Zero- and Few-shot Named Entity Recognition and Text Expansion in Medication Prescriptions using ChatGPT},
pdfsubject={q-bio.NC, q-bio.QM},
pdfauthor={Natthanaphop Isaradech, Andrea Riedel, Wachiranun Sirikul, Markus Kreuzthaler, Stefan Schulz},
pdfkeywords={Named-entity Recognition, Medication Extraction, Large Language Models, Few-shot Learning, Zero-shot Learning, Prompt Engineering},
}

\begin{document}
\maketitle

\begin{abstract}
\textbf{Introduction}: Medication prescriptions are often in free text and include a mix of two languages, local brand names, and a wide range of idiosyncratic formats and abbreviations. Large language models (LLMs) have shown promising ability to generate text in response to input prompts. We use ChatGPT 3.5 to automatically structure and expand medication statements in discharge summaries and thus make them easier to interpret for people and machines.
\textbf{Methods}: Named-entity Recognition (NER) and Text Expansion (EX) are used in a zero- and few-shot setting with different prompt strategies. 100 medication statements were manually annotated and curated. NER performance was measured by using strict and partial matching. For the task EX, two experts interpreted the results by assessing semantic equivalence between original and expanded statements. The model performance was measured by precision, recall, and F1 score.
\textbf{Results}: For NER, the best-performing prompt reached an average F1 score of 0.94 in the test set. For EX, the few-shot prompt showed superior performance among other prompts, with an average F1 score of 0.87.
\textbf{Conclusion}: Our study demonstrates good performance for NER and EX tasks in free-text medication statements using ChatGPT. Compared to a zero-shot baseline, a few-shot approach prevented the system from hallucinating, which would be unacceptable when processing safety-relevant medication data.
\end{abstract}



\section{Introduction}
\label{sec:Introduction}
Prescribing drugs has a high impact on patient safety. In many places, handwritten prescriptions are still common, and only when a discharge summary is written, medication information is registered in electronic health records (EHRs). Here, physicians tend to use compact language and particularly abbreviations. They mix up brands with ingredient names and skip units ("ASA 100") and dose form information (``tablet'', ``suspension'', ``eyedrops''). For route of administration, medication frequency, and time patterns, a broad range of different styles is used. Abbreviated Latin terms such as ``tds'' (three times a day) and ``qPM'' (once in the evening) are common in some jurisdictions, but completely unknown in others where ``1-1-1'' corresponds to ``tds'', and ``0-0-1'' to ``qPM''. Wherever drug prescriptions are done in narrative form, their interpretation - particularly beyond their narrow context of use - remains a challenge. Ideally, a cryptic ``ASA 100 qPM'' should be automatically transformed into a clear ``Acetylsalicylic acid 100 milligrams oral tablet, to be taken orally every afternoon''.

English is the lingua franca in scholarly communications, but EHRs mostly use the official languages of the respective jurisdiction. Yet there are exceptions, such as in Arabic or Asian countries, which have not developed sophisticated medical terminology in their languages to an extent that would suffice for clinical documentation and therefore use English. International healthcare teams, especially in the Middle East, communicate in English. English is the interlingua of multilingual countries such as India and Malaysia. It is therefore unsurprising that many flavors of non-standard medical English have developed, which present challenges both for native and second-language speakers. Beyond affecting communication, we expect medical content in "Asian Englishes" to pose also difficulties for machine processing of medical language, because tools and resources have been trained with mainstream English EHR content and with publications polished and standardized by editors. In a similar vein, controlled vocabularies such as ICD-10 or SNOMED CT do not account for non-mainstream English term variants.

This paper focuses on healthcare documentation in Thailand, which has undertaken a concerted effort to stimulate computerized physician order entry (CPOE) to achieve high-quality prescriptions~\cite{theera-ampornpunt_thai_2011} Nevertheless, most hospitals still depend on written prescriptions, which subsequently require manual input into EHRs \cite{prakob_retrospective_2023, sanguansak_impact_2012}. Narrative medication statements are still used for drug reconciliation and communication \cite{chiewchantanakit_effectiveness_2020}. Such expressions in ``Thai English'' blend two languages and character sets, and use a variety of styles and formats depending on the physician's preferences, such as the merging of trade names and ingredients \cite{prakob_retrospective_2023, noauthor_rational_nodate}, the use of abbreviated names and routes6, and the omission of dosage forms \cite{salmasi_medication_2015}. E.g., in ``Thyrosit (50) 0.5x1 o ac \includegraphics[scale=0.7]{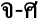}'', ``Thyrosit'' is a trade name for levothyroxine, and ``(50)'' indicates the strength. The unit (here micrograms) is omitted as there is no levothyroxine preparation at the milligram level. ``0.5x1'' indicates the action of consuming half (``0.5'') of the dosage once (``x1'') each day. Finally, ``o ac \includegraphics[scale=0.7]{figures/thaitext.png}'', comprises a mixture of Thai, English, and Latin abbreviations denoting routes and timings. ``o'' represents ``per os (po)'' (by mouth) and ``ac'' means ``ante cibum'' (before meals). \includegraphics[scale=0.7]{figures/thaitext.png} is a Thai abbreviation for ``Monday to Friday''. It is foreseeable that considerable problems arise whenever we aim to automatically extract information from such free-text instructions, although over the last decade machine learning-based approaches have gained popularity in clinical information extraction thanks to the ever-improving performance of human language technologies \cite{hahn_medical_2020}. Large-language models (LLMs) have attracted immense interest after the release of ChatGPT in 2022~\cite{jin_large_2023, dave_chatgpt_2023, vaswani_attention_2023}. Their capacity for handling and analyzing large-scale text and generating content in response to input prompts makes them highly promising for a wide range of applications, from clinical name entity recognition (NER) \cite{ramachandran_extracting_2023, sivarajkumar_healthprompt_2022} to encoding clinical knowledge \cite{singhal_large_2023}. Moreover, LLMs have vastly benefited from transfer learning where their pre-existing knowledge is leveraged and fine-tuned with minimal labeled data or instructions, making them highly adaptable. This contributed to the popularity of zero to few-shot learning paradigms: for each new class, one-shot learning uses just one labeled example, few-shot learning uses a limited set thereof, whereas in zero-shot learning no labeled data are provided at all \cite{kadam_review_2020}. The capabilities of LLMs can be customized, improved, or refined by a set of instructions called a prompt \cite{zhou_large_2023}. To effectively communicate with LLMs, prompt engineering strategies have turned out to be fundamental \cite{white_prompt_2023}. 

Against this background, we focus on the idiosyncratic and compact nature of medication statements in Thai medical records and formulate our research questions as follows:
\begin{itemize}
    \item To which extent is ChatGPT 3.5 able to restructure and normalize medication statements?
    \item Does this depend on prompting patterns?
    \item What are the weaknesses of this approach regarding patient safety?
\end{itemize}

\begin{figure}[h!tbp]
    \centering
    \includegraphics[width=1\linewidth]{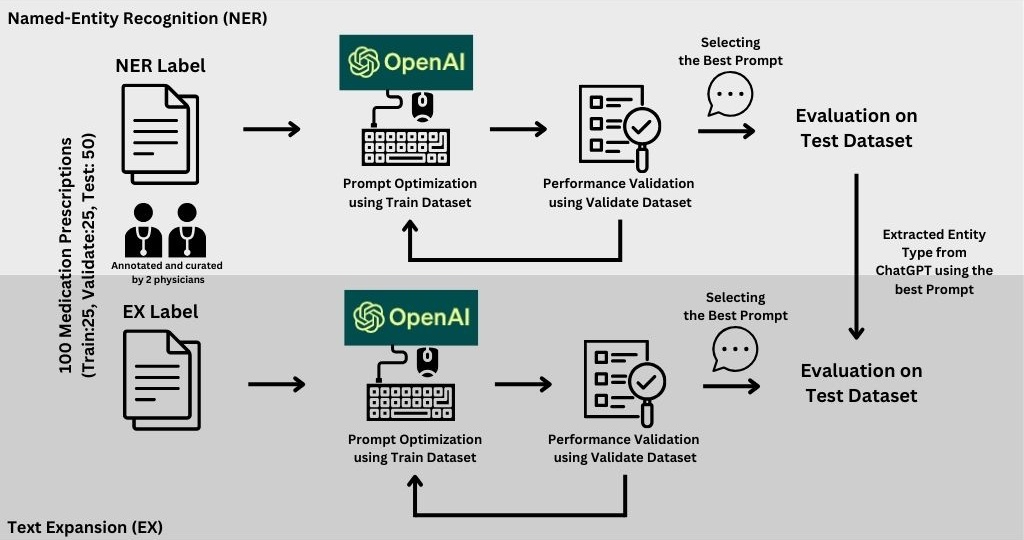}
    \caption{An illustrated overview of Zero- and Few-shot NER (named entity recognition) and EX (text expansion) tasks in our study.}
    \label{fig:overview}
\end{figure}

\section{Methods}
\label{sec:methods}

\subsection{Overview}
\label{subsec:overview}
Our study consists of a named entity recognition (NER) and a text expansion (EX) phase, as shown in Figure~\ref{fig:overview}. The initial step involves the human annotation of medication statements with the entity types Medication, Strength, Unit, Mode, and Instructions, as well as with manually expanded text, done by a domain expert. Prompts are then optimized by utilizing the NER-annotated training dataset (see below). Following this, prompt validation and selection occur on a separate dataset. The selected prompt is then evaluated on a designated test dataset. In the same way, leveraging the outcomes derived from NER prompts, we design and optimize another set of prompts for the task EX, using EX-annotated training data. Finally, prompts find the best-performing one for EX and assess its performance on the test dataset.

\subsection{Dataset}
\label{subsec:dataset}
Nineteen discharge summaries were randomly selected from 6000 ones from the internal medicine department at Maharaj Nakorn Chiang Mai Hospital (2018–2022), authored by internists, residents, or 6th-year medical students. 100 medication prescriptions were manually extracted, annotated by a physician, and curated by another physician. Data was then randomly divided into a training set, a validation set, and a test set with a 25:25:50 split as shown in Table~\ref{subsec:dataset}.

\subsection{NER Annotation}
\label{subsec:entityTypesForNER}

\begin{table}[h!tbp]
    \centering
    \begin{tabular}{lccccc}
    \toprule
                \textbf{Dataset} &\textbf{Medication} & \textbf{Strength} & \textbf{Unit} & \textbf{Mode} & \textbf{Instructions} \\
    \midrule
    Training   & 25                    & 20                 & 21                & 23            & 25                               \\
    Validation & 25                    & 19                 & 19                & 22            & 24                               \\
    Test       & 50                   & 38                 & 40               & 43           & 50                              \\
    \bottomrule
    \end{tabular}
    \caption{Number of Instances per entity type in the Training, Validation, and Test data.}
    \label{dataset}
\end{table}

The annotations were divided into two separate sections, one for NER annotation and one for EX annotation. For NER, the following five entity types were distinguished:

\begin{description}
    \item[Medication.] A brand, a substance, or a set of substances that make up a single drug product, depending on how it was mentioned in the prescription. Example: ``Paracetamol'', ``ASA'', ``Thyrosit'', ``Sulfamethoxazole/Trimethoprim''.
    \item[Strength.] The amount of the substance. Given a prescription, "Paracetamol (500) po q 4-6 hr", "500" represents the strength of the substance, and the unit "milligram" is omitted.
    \item[Unit.] The unit in which the strength is measured, such as milligrams (``mg'') or micrograms (``mcg''). The unit is often omitted, but can be inferred from the context.
    \item[Mode.] The route for administering the medication, e.g. ``po'' (per oral), ``opc'' (per oral and postprandial).
    \item[Instructions.] A narrative explanation of how to take the medication including Quantity of Dose Form, Dose Form, Relation to Meal, Frequency, and Other. For example, "po pc q 4-6 hr" means consuming the medication by oral route (po; per oral), after meal (pc; postpandrial), and do it every 4-6 hours (q 4-6 hr.)
\end{description}

\subsection{EX Annotation}
\label{subsec:exAnnotationExpansionTypes}
For the EX annotation, the expansion types of the text were classified into eight categories. The latter five of them, viz. Quantity of Dose Form, Dose Form, Relation to Meal, Frequency, and Others extend the \textit{Instructions} entity type:

\begin{description}
    \item [Active ingredients.] Substituting the text in the Medication entity type by the corresponding active ingredient(s). E.g., converting ``paracet'' to ``paracetamol'' or ``Thyrosit'' to ``levothyroxine''.
    \item [Unit.] Expanding the short form Unit to a long-form, e.g. ``milligram'' for ``m.g.'' and ``mg''.
    \item [Mode.] Expanding the Mode entity type into a non-abbreviated description of the route of administration, such as ``po'' to ``oral''.
    \item [\textit{Instructions} Dose Form.] Describing the form of drug dose, such as tablet, inhaler, or droplet.
    \item [\textit{Instructions} Quantity of Dose Form.] Expanding information related to the quantity associated with the Dose Form expansion type, e.g., in ``Take Simvastatin 40 mg 0.5 tablet once daily'', ``0.5'' indicates Quantity of Dose Form, i.e., without using abbreviations or ambiguous terms.
    \item [\textit{Instructions} Relation to Meal.] Describing the relation of the medication intake related to meals from the \textit{Instructions} entity type, including ``before meals'', ``after meals'', and ``before bed''.
    \item [\textit{Instructions} Frequency.] Expanding details regarding how often the medication should be taken, e.g. ``once daily'', ``once weekly'', or ``twice daily''.
    \item [\textit{Instructions} Other.] Capturing any additional information in \textit{Instructions} that do not fit into the EX categories, such as the duration of the prescription, or conditional instructions such as ``take when experiencing palpitations''.
\end{description}

\subsection{Prompt engineering}
\label{subsec:promptEngineering}
Our approach to zero-shot learning prompt design was based on combinations of prompt patterns, as recently suggested \cite{white_prompt_2023}. We first created prompts using the training dataset and assessed the ChatGPT outcomes by applying these prompts to the validation set. Then we refined the prompt combinations to enhance the outcome performance, guided by the results from the validation set. Once we achieved satisfactory results, we selected the most effective prompt based on the validation set and evaluated it on the test dataset. For the NER task, we selected the following patterns for our prompt design:

\begin{description}
    \item [Persona.] Assigning a role provides context and direction, enabling the LLM to adopt a specific identity and tailor its responses accordingly.
    \item [Template.] Assigning the LLM a precise template to adjust its output to a specific format.
    \item [Few-shot.] Assigning the LLM prescription examples from the annotated training dataset for the LLM to understand what the output looks like. In our case, we simply give the example of annotated NER in tabular format.
\end{description}

The prompts were structured based on combinations of patterns. 

\subsubsection{NER Prompt Patterns}
For the Named Entity Recognition (NER) task, six different prompt combinations are considered:

\begin{description}
    \item[Prompt A.] This prompt combines \textbf{Persona} and \textbf{Template} patterns.
    \item[Prompt B.] This prompt uses only the \textbf{Template} pattern.
    \item[Prompt C.] This prompt uses the \textbf{Template} pattern along with \textbf{five} examples of correct NER recognition.
    \item[Prompt D.] This prompt combines the \textbf{Persona} and \textbf{Template} patterns, including \textbf{five} examples of correct NER recognition.
    \item[Prompt E.] This prompt uses the \textbf{Template} pattern with \textbf{ten} examples of correct NER recognition.
    \item[Prompt F.] This prompt combines the \textbf{Persona} and \textbf{Template} patterns with \textbf{ten} examples of correct NER recognition.
\end{description}

\subsubsection{EX Prompt Patterns}
For the EX task, three different prompt combinations are considered:

\begin{description}
    \item[Prompt 1.] This is a plain prompt command that does not follow any specific prompt pattern.
    \item[Prompt 2.] This prompt combines \textbf{Persona} and \textbf{Template} patterns.
    \item[Prompt 3.] This prompt combines \textbf{Persona} and \textbf{Template} patterns along with \textbf{five} examples of correct NER recognition.
\end{description}

The details of the prompts are in Appendices \ref{appendix: prompt-ner} and \ref{appendix: prompt-ex}.

To gather NER and EX results, we manually collected the ChatGPT outputs and organized the NER and EX results in a spreadsheet. For every prompt, a new thread was created to keep ChatGPT's responses independent. Where no response was generated, we left the corresponding cell empty.

\subsection{Evaluation}
\label{subsec:evaluation}
Precision (P), recall (R), and averaged F1 scores were used to evaluate the model performance. For NER, scores were computed using strict and partial matching criteria. Strict matching required identical spans and a correct entity type. Partial matching only required overlapping of spans and correct entity types. The evaluation of the EX task was based on inspection by two domain experts. They carefully reviewed each result and assessed where semantic equivalence could be asserted, according to their interpretation of the intended meaning of the prescription statement in the context of their expertise.

\begin{figure}[h!tbp]
    \centering
    \includegraphics[width=1\linewidth]{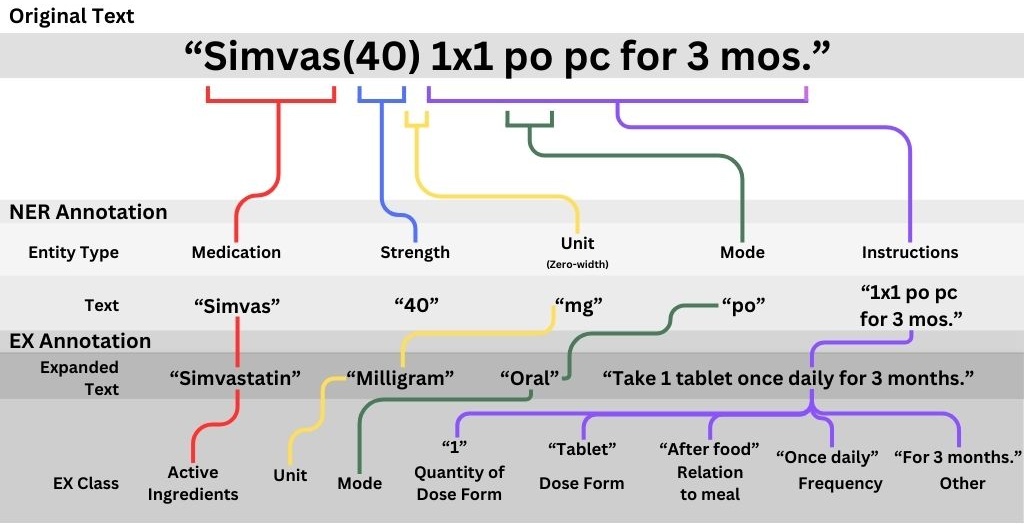}
    \caption{Typical medication statement, its components, and annotations, including zero-width annotations for ellipses for which the meaning can be inferred from the context, such as "mg" in this example.}
    \label{annotation}
\end{figure}

Figure~\ref{annotation} depicts how we analyze text. For the original ``Simvas(40) 1x1 po pc for 3 mos.,'' our NER annotation labels spans and entity types at the token level, excluding the special characters ``()[]x*''. In the EX phase, annotators expand the components Active ingredients, Units, Mode, and Instructions. For Instructions, annotators expand terms based on five classes: Quantity of dose form, Dose form, Relation to Meal, Frequency, and Other.

In the evaluation phase, ChatGPT’s generated NER output is compared with our annotations using strict and partial matching. E.g., if ``Simvas(40),'' is identified as medication, it would be false for strict matching and true for partial matching (because the strength ``(40)'' is a separate entity type). Regarding ChatGPT's EX output, each expanded class is compared for semantic equivalence. A response ``by mouth'' instead of the gold standard annotation ``oral'' is considered true because the meaning is judged identical by the experts. For Instructions, five additional categories assess the output quality. Evaluators carefully examine each category by comparing the expanded Instructions against the gold standard. E.g., if ChatGPT expanded ``take 1 tablet once daily for 3 months after a meal'', we would compare ``1'', ``tablet'', ``once daily'', ``after a meal'', and ``for three months'' with the gold standard \textit{Quantity of dose form, Dose form, Frequency, Relation to Meal, and Other}, respectively. 

\section{Results}
\label{sec:results}

\subsection{Zero- and few-shot named-entity recognition}
\label{subsec:zeroAndFewShotNER}

The NER results are shown in Table \ref{partial-matching} and \ref{strict-matching}. Prompts A and B showed good performance with an average F1 of 0.72 and 0.74, respectively, based on partial matching. Prompts C, D, E, and F demonstrated better overall performance than A or B. However, even though prompt C generally had a better overall F1 than prompt A, it exhibited the poorest Unit detection compared to all prompts, even with the same prompt pattern and more examples. Prompt D showed a similar issue, as it failed to recognize Instructions effectively. Prompts E and F showed superior performance compared to the others, with an average F1 of 0.94 and 0.90, respectively (partial matching). Both prompts revealed promising for medical NER tasks. Nevertheless, we decided to prioritize prompt E, as it had the highest average F1 and better NER performance on Unit in partial matching.

\begin{table}[h!tbp]
    \centering
    \caption{Performance Metrics Across Different Prompts for Partial Matching}
        \begin{tabular}{lcccccc}
        \toprule
                  & \textbf{Prompt A} & \textbf{Prompt B} & \textbf{Prompt C} & \textbf{Prompt D} & \textbf{Prompt E} &  \textbf{Prompt F} \\ 
        \midrule
        \textbf{Medication} & & & & & & \\ 
        Precision & 1.00 & 1.00 & 1.00 & 1.00 & 1.00 & 1.00 \\ 
        Recall    & 1.00 & 1.00 & 1.00 & 0.96 & 1.00 & 1.00 \\ 
        F1 score  & 1.00 & 1.00 & 1.00 & 0.98 & 1.00 & 1.00 \\
        \midrule
        \textbf{Strength} & & & & & & \\ 
        Precision & 1.00 & 1.00 & 1.00 & 1.00 & 1.00 & 1.00 \\ 
        Recall    & 1.00 & 0.96 & 1.00 & 1.00 & 1.00 & 1.00 \\ 
        F1 score  & 1.00 & 0.98 & 1.00 & 1.00 & 1.00 & 1.00 \\
        \midrule
        \textbf{Unit} & & & & & & \\ 
        Precision & 1.00 & 1.00 & 1.00 & 1.00 & 1.00 & 1.00 \\ 
        Recall    & 0.64 & 0.68 & 0.32 & 0.68 & 0.40 & 0.32 \\ 
        F1 score  & 0.78 & 0.81 & 0.48 & 0.81 & 0.57 & 0.48 \\
        \midrule
        \textbf{Mode} & & & & & & \\ 
        Precision & 0.96 & 1.00 & 1.00 & 1.00 & 1.00 & 1.00 \\ 
        Recall    & 0.96 & 1.00 & 1.00 & 1.00 & 1.00 & 1.00 \\ 
        F1 score  & 0.96 & 1.00 & 1.00 & 1.00 & 1.00 & 1.00 \\
        \midrule
        \textbf{Instructions} & & & & & & \\ 
        Precision & 0.00 & 0.00 & 0.60 & 1.00 & 1.00 & 1.00 \\ 
        Recall    & 0.00 & 0.00 & 0.60 & 0.68 & 1.00 & 1.00 \\ 
        F1 score  & 0.00 & 0.00 & 0.60 & 0.81 & 1.00 & 1.00 \\
        \midrule
        \textbf{Avg Precision} & 0.79 & 0.80 & 0.92 & 1.00 & 1.00 & 1.00 \\ 
        \textbf{Avg Recall}    & 0.66 & 0.69 & 0.74 & 0.74 & 0.80 & 0.82 \\ 
        \textbf{Avg F1 score}  & 0.72 & 0.74 & 0.78 & 0.82 & \textbf{0.94} & 0.90 \\
        \bottomrule
        \end{tabular}
    
    \label{partial-matching}
\end{table}

\begin{table}[h!tbp]
    \caption{Medical Entity Recognition Performance using different ChatGPT 3.5 Prompts on the validation set (n = 25). 0.95 confidence intervals for typical values: 1.00 [0.86, 1.00]; 0.80 [0.59, 0.93]; 0.40 [0.21, 0.61].}
    \centering
        \begin{tabular}{lcccccc}
        \toprule
                  & \textbf{Prompt A} & \textbf{Prompt B} & \textbf{Prompt C} & \textbf{Prompt D} & \textbf{Prompt E} & \textbf{Prompt F} \\ 
        \midrule
        \textbf{Medication} & & & & & & \\ 
        Precision & 0.72 & 0.76 & 0.80 & 0.72 & 0.80 & 0.80 \\ 
        Recall    & 0.72 & 0.76 & 0.80 & 0.72 & 0.80 & 0.80 \\ 
        F1 score  & 0.72 & 0.76 & 0.80 & 0.72 & 0.80 & 0.80 \\
        \midrule
        \textbf{Strength} & & & & & & \\ 
        Precision & 1.00 & 0.98 & 1.00 & 1.00 & 1.00 & 1.00 \\ 
        Recall    & 1.00 & 0.96 & 1.00 & 1.00 & 1.00 & 1.00 \\ 
        F1 score  & 1.00 & 0.97 & 1.00 & 1.00 & 1.00 & 1.00 \\
        \midrule
        \textbf{Unit} & & & & & & \\ 
        Precision & 0.78 & 0.81 & 0.74 & 0.86 & 0.75 & 0.76 \\ 
        Recall    & 0.60 & 0.72 & 0.32 & 0.84 & 0.40 & 0.52 \\ 
        F1 score  & 0.68 & 0.76 & 0.45 & 0.85 & 0.52 & 0.62 \\
        \midrule
        \textbf{Mode} & & & & & & \\ 
        Precision & 0.96 & 0.96 & 0.96 & 0.92 & 0.96 & 0.96 \\ 
        Recall    & 0.96 & 0.96 & 0.96 & 0.92 & 0.96 & 0.96 \\ 
        F1 score  & 0.96 & 0.96 & 0.96 & 0.92 & 0.96 & 0.96 \\
        \midrule
        \textbf{Instructions} & & & & & & \\ 
        Precision & 0.00 & 0.00 & 0.60 & 0.20 & 0.80 & 0.84 \\ 
        Recall    & 0.04 & 0.04 & 0.60 & 0.24 & 0.80 & 0.84 \\ 
        F1 score  & 0.00 & 0.00 & 0.60 & 0.22 & 0.80 & 0.84 \\
        \midrule
        \textbf{Avg Precision} & 0.69 & 0.70 & 0.82 & 0.74 & 0.86 & 0.87 \\ 
        \textbf{Avg Recall}    & 0.66 & 0.69 & 0.74 & 0.74 & 0.80 & 0.82 \\ 
        \textbf{Avg F1 score}  & 0.67 & 0.70 & 0.78 & 0.74 & 0.83 & \textbf{0.84} \\
        \bottomrule
        \end{tabular}
    
    \label{strict-matching}
\end{table}

The results of the NER task in the test dataset are shown in Table \ref{ner-test-transposed}. Overall, ChatGPT commanded by prompt E, generated good performance for the NER task with an average F1 score for all entity types of 0.79 and 0.92 based on strict and partial matching, respectively. However, the test dataset findings highlighted that the Unit continues to be the poorest-performing in NER. Compared to the validation dataset, the performance metrics for the test dataset are slightly inferior, indicating that ChatGPT has a high generalizability for identifying entity types.

\begin{table}[h!tbp]
    \caption{Medical Entity Recognition Performance using Prompt E on ChatGPT 3.5 evaluated with Strict and Partial Matching - Test set (n=50). 0.95 confidence intervals for typical values: 1.00 [0.93, 1.00]; 0.78 [0.64, 0.88]; 0.38 [0.25, 0.53]}
    \centering
        \begin{tabular}{lcccccc}
        \toprule
        & \textbf{Precision$_{St}$} & \textbf{Recall$_{St}$} & \textbf{F1$_{St}$} & \textbf{Precision$_{Pa}$} & \textbf{Recall$_{Pa}$} & \textbf{F1$_{Pa}$} \\
        \midrule
        Medication   & 0.80 & 0.78 & 0.79 & 1.00 & 0.98 & 0.99 \\ 
        Strength     & 0.96 & 0.98 & 0.97 & 1.00 & 0.98 & 0.99 \\
        Unit         & 0.81 & 0.38 & 0.52 & 1.00 & 0.34 & 0.51 \\
        Mode         & 0.95 & 0.96 & 0.95 & 1.00 & 1.00 & 1.00 \\
        Instructions & 0.67 & 0.66 & 0.67 & 1.00 & 0.98 & 0.99 \\
        \midrule
        Average     & 0.84 & 0.75 & 0.79 & 1.00 & 0.86 & 0.92 \\
        \bottomrule
        \end{tabular}
    \label{ner-test-transposed}
\end{table}

\subsection{Zero- and few-shot text expansion and manual assessment}
\label{subsec:zeroAndFewShotNER}

After obtaining metrics from NER prompts, we passed the ChatGPT output of the best-performing prompt E to the EX phase of our experiment. The NER output was converted to a tabular format to put the text into the ChatGPT chatbox. The EX performance results are shown in Table \ref{text-expansion-results}. In the validation dataset, Prompt 3 had the best overall performance on term expansion in all categories, excluding Unit, with an average F1 score of 0.87. Prompts 2 and 3 were generally better as they were given more examples of medication prescriptions. Prompt 1, being the simplest EX prompt, was not capable of expanding any terms in the Mode class at all. Prompt 3 was then selected for test dataset evaluation, and the results showed an average F1 of 0.77, which decreased by 0.10 compared to the validation set because the metrics for Unit and [Instructions] Relation to Meal dropped largely compared to the validation dataset.

\begin{table}[h!tbp]
    \caption{Text Expansion Results on Validation set (n = 25) and Test Dataset (n = 50). See previous tables for typical confidence intervals.}
    \centering
        \begin{tabular}{lcccc}
        \toprule
                  & \textbf{Validation} & & & {\textbf{Test}} \\ 
                  & \textbf{Prompt 1} & \textbf{Prompt 2} & \textbf{Prompt 3} & \textbf{Prompt 3} \\ 
        \midrule
        \textbf{Active Ingredient} & & & & \\ 
        Precision & 1.00 & 1.00 & 1.00 & 1.00 \\ 
        Recall    & 0.64 & 0.60 & 0.68 & 0.60 \\ 
        F1 score  & 0.78 & 0.81 & 0.84 & 0.75 \\
        \midrule
        \textbf{Unit} & & & & \\ 
        Precision & 0.06 & 0.84 & 0.83 & 0.04 \\ 
        Recall    & 0.24 & 0.48 & 0.44 & 0.20 \\ 
        F1 score  & 0.10 & 0.61 & 0.58 & 0.07 \\
        \midrule
        \textbf{Mode} & & & & \\ 
        Precision & 0.00 & 1.00 & 1.00 & 1.00 \\ 
        Recall    & 0.00 & 0.88 & 0.84 & 0.80 \\ 
        F1 score  & 0.00 & 0.94 & 0.91 & 0.89 \\
        \midrule
        \textbf{[Instructions]} \textbf{Quantity of Dose Form} & & & & \\ 
        Precision & 1.00 & 1.00 & 1.00 & 0.98 \\ 
        Recall    & 0.92 & 0.96 & 0.92 & 0.94 \\ 
        F1 score  & 0.96 & 0.98 & 0.96 & 0.96 \\
        \midrule
        \textbf{[Instructions]} \textbf{Dose Form} & & & & \\ 
        Precision & 1.00 & 0.89 & 0.98 & 0.96 \\ 
        Recall    & 0.96 & 0.48 & 0.94 & 0.94 \\ 
        F1 score  & 0.98 & 0.62 & 0.96 & 0.95 \\
        \midrule
        \textbf{[Instructions]} \textbf{Relation to Meal} & & & & \\ 
        Precision & 0.88 & 0.87 & 0.94 & 0.73 \\ 
        Recall    & 0.80 & 0.76 & 0.92 & 0.42 \\ 
        F1 score  & 0.84 & 0.81 & 0.92 & 0.53 \\
        \midrule
        \textbf{[Instructions]} \textbf{Frequency} & & & & \\ 
        Precision & 0.94 & 0.93 & 0.93 & 0.91 \\ 
        Recall    & 0.80 & 0.06 & 0.60 & 0.90 \\ 
        F1 score  & 0.86 & 0.11 & 0.73 & 0.91 \\
        \midrule
        \textbf{[Instructions]} \textbf{Others} & & & & \\ 
        Precision & 1.00 & 1.00 & 1.00 & 0.96 \\ 
        Recall    & 1.00 & 1.00 & 1.00 & 0.96 \\ 
        F1 score  & 1.00 & 1.00 & 1.00 & 0.96 \\
        \midrule
        \textbf{Average Precision} & 0.74 & 0.94 & 0.96 & 0.82 \\ 
        \textbf{Average Recall}    & 0.67 & 0.65 & 0.79 & 0.72 \\ 
        \textbf{Average F1 score}  & 0.70 & 0.77 & \textbf{0.87} & 0.77 \\
        \bottomrule
        \end{tabular}
    \label{text-expansion-results}
\end{table}

\section{Discussion}
\label{sec:discussion}

\subsection{Comparison to previous work}

Early attempts to use ChatGPT for clinical NLP were made by Hu et al., but its performance lagged behind that of the supervised BioClinicalBERT model, regarding the extraction of treatments mentioned in discharge summaries \cite{hu_zero-shot_2023}. However, they did not distinguish drug treatments from other medical treatments. Therefore, it seems that our work is the first to use ChatGPT for specific and detailed medication statement analysis.

Comparing our NER results (mean F1 score 0.92) to an earlier study (mean F1 score 0.80) using CRF shows a clear superiority \cite{tao_prescription_2017}. In that study, medication details were improperly delineated whenever punctuation patterns such as brackets did not follow the norm \cite{wang_future_2023}. However, problems like this can mostly be seen as resolved by deep learning. So did the unusual bracketing of strength (e.g., "ASA (500)") not constitute any obstacle.

More interesting is the comparison with MT-NER-PMB (2021), which can be seen as a recent non-LLM baseline, which outperformed other BERT models. Comparing our results to their F1 score for partial results we could show slightly better detections for Medication (0.99 vs 0.96), Mode (1.00 vs. 0.95), and Strength (0.99 vs. 0.98). However, our study separates Strength and Unit, therefore the relatively low F1 score for Unit 0.51 is not integrated in the calculation for Strength \cite{narayanan_contextual_2022}.

That our medication statements often did not contain units of measurement (in the gold standard annotation constituting so-called zero-with annotations, cf. Fig. \ref{annotation}) led to a low precision in Prompt 1 because it repeated the NER task by adding the found abbreviation in the text. The results suggest that the addition of prompt patterns in prompts 2 and 3 solved this.

Compared to traditional machine learning, it was astonishing to observe how few training examples produced a considerable benefit. This highlights the importance of good prompt engineering to optimally exploit the content of an LLM. However, we cannot explain the poor result for Unit in NER, whereas the even worse performance of Unit in EX (such as from "mg" to "Milligram") may result from the fact that unit symbols are hardly ever expanded, even in scholarly or popular publications or drug leaflets.

\subsection{Error Analysis}

An important result of our study was that incremental LLM prompting with repetitive human assessment improved the result and the NER task achieved an F1 of 0.79 and 0.92 on the test set for strict and partial matching, respectively, with the recognition of units of measurement (which were often omitted) showing the poorest results. The Unit expansion also performed poorly in the EX task, which achieved an F1 of 0.07 only, against 0.77 for all tasks. The expansion of abbreviated drug names and brand names to active ingredients achieved a 1.00 precision such as the expansion of mode of administration, which shows that the result has in no way deteriorated compared to the original text. It is important to highlight that from a patient safety point of view, precision is more important than recall. A low recall means that the system did not improve the quality of the medication statement, whereas any suboptimal precision value bears the risk that the content of the statement is distorted.

Unclear short forms are a well-known problem in clinical texts \cite{kreuzthaler_unsupervised_2016}. Abbreviation of drugs often follows local jargon. Interestingly, only one ("ASA") was correctly resolved, whereas the other four were not resolved at all ("MFM" for metformin, "MTV" for multivitamins, and "K" for potassium). Regarding ChatGPT’s risk of hallucination, it is remarkable that the system opted for non-resolution instead of for wrong resolution. While in prompt D the abbreviation "ORS" for "oral rehydration solution" was even not detected during the NER validation phase as Medication, in prompt E at least in NER it was detected in validation and test but not extended by test EX. Despite the fact it is a common abbreviation in our data set, the complexity increased due to a mixture of languages and documentation styles. Our results for EX show a drastic increase in F1 for the expansion type Mode from Prompt 1 without any prompt patterns (F1 score: 0.00) to Prompt 2 with adding Persona and Template (F1 score 0.94). In Prompt 2, ChatGPT detected all abbreviations of "oral" in all samples, but still tried to translate each different abbreviation literally like "opc" to "oral, by consumption" or "po" "per mouth". By giving 5 examples with the prompt pattern "Few-shot" in Prompt 3, ChatGPT detected, e.g., "po", "o", "opc", and "po ac" as "oral". Even "oac", which was not given as an example, was now correctly defined as "oral" and not as "oral administration". The changed definition of Mode leads to an increase of F1 by 0.11 for the expansion type Relation to Meal in Prompt 3 as well, because ChatGPT can now categorize the subsequent information. In contrast to the quality improvement of ChatGPT’s output for Mode, its F1 slightly decreased by 0.03. Adding clearly defined case-specific prompt patterns for foreign language abbreviations is an essential step in creating the expected results.

Even though most of our medication statements are in English, some Thai brand names such as "HIDIL" show the non-English vocabulary limitations of ChatGPT. For the expansion type Active Ingredients, the main positive results are the already detected active ingredients in the NER phase in comparison to medication brands, which are less often transformed to the active ingredient, e.g. instead of recognizing "HIDIL" as Gemfibrozil, ChatGPT inserted the brand name "HIDIL". This parallels a similar finding when querying ChatGPT with the Chinese medication "Motrin" (ibuprofen) which was mistakenly linked to aspirin when retrieving adverse drug reactions-related information \cite{wang_future_2023}.

\subsection{Limitations}

Hallucinations are a relevant issue in ChatGPT \cite{fink_potential_2023, lee_utilizing_2023}. Here, we obtained the remarkable result one instance of a hallucinated response could be identified in the test data (n=50). The daily dosage instruction of an antipsychotic drug was interpreted correctly, but ChatGPT added an "as needed" statement on top, however, regarding a small dose that is not expected to cause harm.

Given the original text prescription, “Ridperidone (1) 0.5x1 po hs”, ChatGPT gave us “0.5 Tablet oral at bedtime; 0.25 Tablet oral as needed for agitation” for instructions EX task instead of “0.5 Tablet Oral before bed Once daily” which is the correct answer.

The size of the dataset must also be mentioned. Particularly, the assessment of rare but important outcomes such as the occurrence and the severity of hallucinations would have required much more data. Another limitation was the restriction to ChatGPT 3.5. We could have compared it to other LLMs, but our focus has been on the comparison of different prompting strategies rather than different models.

\section{Conclusions and Outlook}
\label{sec:conclusionAndOutlook}
Converting medication information into a standardized format is expected to improve patient safety by making it easier to interpret by both humans and machines. This study demonstrated good performance in the task of entity recognition (NER) and entity expansion (EX) in free-text medication statements using ChatGPT 3.5, in comparison to related work. The data, taken from Thai medical records, exhibited particularities in style and format and used a mixture of English and Thai, including local drug product names. We were able to demonstrate good performance in NER and EX. Compared to a zero-shot baseline, a 10-shot approach prevented the system from hallucinating, which would be unacceptable when processing highly safety-relevant medication data.

Our study works on anonymized text snippets of real-world data. Further tests and analysis of huge real-world patient data for training are necessary to use ChatGPT on a variety of disciplines and text document types. In the future for the use of real-world patient data, especially the ethical and data safety issues need to be discussed \cite{sallam_chatgpt_2023, beltrami_consulting_2023}. A good alternative would be an LLM deployment on-premise. Like Hu et al. with ChatGPT version 3, we experienced "a significant degree of randomness" while performing the equal prompts more often with version 3.5 as well. In our study, the input sequence length was even longer than the one presented in the publication of Hu et al \cite{hu_zero-shot_2023}.

As we are currently at the very beginning of the LLM era, we expect even better performance in the future. Medication-related statements are a particularly challenging use case because their overly compact and idiosyncratic style constitutes a risk for human and machine misinterpretation with unforeseeable impacts on patient safety on the one hand. On the other hand, a similar risk may occur from even minor hallucinations and imprecisions when expanded and normalized by AI methods. Finally, medication styles drastically vary across languages and jurisdictions. We therefore advocate investing particular efforts in the creation of multilingual, multicultural, and multi-specialty medication benchmarks.

\section*{Data access}
The dataset was approved by the Research Ethics Committee, Faculty of Medicine, Chiang Mai University, Thailand: COM-256509305, Research-ID: 9305.

\section*{Acknowledgement}
AR’s fellowship was granted by the German BMBF within the DAAD IFI programme. NI’s fellowship was granted by OeAD-GmbH/MPC in cooperation with ASEA-UNINET, financed by the Austrian BMBWF. This manuscript is part of a thesis for NI’s Ph.D. in Digital Health, Faculty of Medicine, Chiang Mai University.
 
\section*{Conflict of Interest}
Nothing to declare.

\appendix

\section{Prompts Used in NER Task for ChatGPT 3.5}
\label{appendix: prompt-ner}

\paragraph{\textbf{Prompt A}}
You are now a Named Entity Recognition Model. I will give you a list of narrative drug prescriptions. Please slice the narrative text based on the Entity Types you detect and organize it as a table record with the columns: Medication ET, Strength ET, Unit ET, Quantity of Dose Form per intake ET, Dose Form ET, Mode ET, Timing ET, Frequency ET, Duration ET, Instructions ET (Dose, Frequency, Duration) without changing anything in the narrative prescription. For missing values, leave the cell blank.

\paragraph{\textbf{Prompt B}}
I will give you a list of narrative drug prescriptions. Please slice the narrative text based on the Entity Types you detect and organize it as a table record with the columns: Medication ET, Strength ET, Unit ET, Quantity of Dose Form per intake ET, Dose Form ET, Mode ET, Timing ET, Frequency ET, Duration ET, Instructions ET (Dose, Frequency, Duration) without changing anything in the narrative prescription. For missing values, leave the cell blank.

\paragraph{\textbf{Prompt C}}
I will give you a list of narrative drug prescriptions. Please slice the narrative text based on the Entity Types you detect and organize it as a table record with the columns: Medication ET, Strength ET, Unit ET, Quantity of Dose Form per intake ET, Dose Form ET, Mode ET, Timing ET, Frequency ET, Duration ET, Instructions ET (Dose, Frequency, Duration) without changing anything in the narrative prescription. For missing values, leave the cell blank.

Here are some examples that you can study with:

\begin{itemize}
    \item Xarator (40) 1/2x1 opc, Xarator, 40, mg, 1/2, tablet, opc, opc, 1/2x1, , 1/2x1 opc
    \item Douzabox 1x2 opc, Douzabox, , , 1, tablet, opc, opc, 1x2, , 1x2 opc
    \item omeprazole (20) 1x1 po ac, omeprazole, 20, mg, 1, tablet, po, ac, 1x1, , 1x1 po ac
    \item Thyrosit (50) 0.5x1 o ac figures/thaitext.png, Thyrosit, 50, mcg, 0.5, tablet, o, ac, 0.5x1, \includegraphics[scale=0.7]{figures/thaitext.png}, 0.5x1 o ac \includegraphics[scale=0.7]{figures/thaitext.png}
    \item Mevalotin Pretect (40) 1x1 o pc, Mevalotin Pretect, 40, mg, 1, tablet, o, pc, 1x1, , 1x1 o pc
\end{itemize}

\paragraph{\textbf{Prompt D}}
You are now a Named Entity Recognition Model. I will give you a list of narrative drug prescriptions. Please slice the narrative text based on the Entity Types you detect and organize it as a table record with the columns: Medication ET, Strength ET, Unit ET, Quantity of Dose Form per intake ET, Dose Form ET, Mode ET, Timing ET, Frequency ET, Duration ET, Instructions ET (Dose, Frequency, Duration) without changing anything in the narrative prescription. For missing values, leave the cell blank.

Here are some examples that you can study with:

\begin{itemize}
    \item Xarator (40) 1/2x1 opc, Xarator, 40, mg, 1/2, tablet, opc, opc, 1/2x1, , 1/2x1 opc
    \item Douzabox 1x2 opc, Douzabox, , , 1, tablet, opc, opc, 1x2, , 1x2 opc
    \item omeprazole (20) 1x1 po ac, omeprazole, 20, mg, 1, tablet, po, ac, 1x1, , 1x1 po ac
    \item Thyrosit (50) 0.5x1 o ac \includegraphics[scale=0.7]{figures/thaitext.png}, Thyrosit, 50, mcg, 0.5, tablet, o, ac, 0.5x1, \includegraphics[scale=0.7]{figures/thaitext.png}, 0.5x1 o ac \includegraphics[scale=0.7]{figures/thaitext.png}
    \item Mevalotin Pretect (40) 1x1 o pc, Mevalotin Pretect, 40, mg, 1, tablet, o, pc, 1x1, , 1x1 o pc
\end{itemize}

\paragraph{\textbf{Prompt E}}
I will give you a list of narrative drug prescriptions. Please slice the narrative text based on the Entity Types you detect and organize it as a table record with the columns: Medication ET, Strength ET, Unit ET, Quantity of Dose Form per intake ET, Dose Form ET, Mode ET, Timing ET, Frequency ET, Duration ET, Instructions ET (Dose, Frequency, Duration) without changing anything in the narrative prescription. For missing values, leave the cell blank.

Here are some examples that you can study with:

\begin{itemize}
    \item Xarator (40) 1/2x1 opc, Xarator, 40, mg, 1/2, tablet, opc, opc, 1/2x1, , 1/2x1 opc
    \item Douzabox 1x2 opc, Douzabox, , , 1, tablet, opc, opc, 1x2, , 1x2 opc
    \item omeprazole (20) 1x1 po ac, omeprazole, 20, mg, 1, tablet, po, ac, 1x1, , 1x1 po ac
    \item Thyrosit (50) 0.5x1 o ac \includegraphics[scale=0.7]{figures/thaitext.png}, Thyrosit, 50, mcg, 0.5, tablet, o, ac, 0.5x1, \includegraphics[scale=0.7]{figures/thaitext.png}, 0.5x1 o ac \includegraphics[scale=0.7]{figures/thaitext.png}
    \item Mevalotin Pretect (40) 1x1 o pc, Mevalotin Pretect, 40, mg, 1, tablet, o, pc, 1x1, , 1x1 o pc
    \item Ridperidone (1) 0.5x1 po hs, Ridperidone, 1, mg, 0.5, tablet, po, hs, 0.5x1, , 0.5x1 po hs
    \item Cotrimazole 2 tab po od, Cotrimazole, , , 2, tablet, po, , od, , 2 tab po od
    \item Trazodone (50) 1x1 po hs, Trazodone, 50, mg, 1, tablet, po, hs, 1x1, , 1x1 po hs
    \item LoRANTA (100) 1x1, LoRANTA, 100, mg, 1, tablet, , , 1x1, , 1x1
    \item Nuelin SR (200) 1x2, Nuelin SR, 200, mg, 1, tablet, , , 1x2, , 1x2
\end{itemize}

\paragraph{\textbf{Prompt F}}
You are now a Named Entity Recognition Model. I will give you a list of narrative drug prescriptions. Please slice the narrative text based on the Entity Types you detect and organize it as a table record with the columns: Medication ET, Strength ET, Unit ET, Quantity of Dose Form per intake ET, Dose Form ET, Mode ET, Timing ET, Frequency ET, Duration ET, Instructions ET (Dose, Frequency, Duration) without changing anything in the narrative prescription. For missing values, leave the cell blank.

Here are some examples that you can study with:

\begin{itemize}
    \item Xarator (40) 1/2x1 opc, Xarator, 40, mg, 1/2, tablet, opc, opc, 1/2x1, , 1/2x1 opc
    \item Douzabox 1x2 opc, Douzabox, , , 1, tablet, opc, opc, 1x2, , 1x2 opc
    \item omeprazole (20) 1x1 po ac, omeprazole, 20, mg, 1, tablet, po, ac, 1x1, , 1x1 po ac
    \item Thyrosit (50) 0.5x1 o ac \includegraphics[scale=0.7]{figures/thaitext.png}, Thyrosit, 50, mcg, 0.5, tablet, o, ac, 0.5x1, \includegraphics[scale=0.7]{figures/thaitext.png}, 0.5x1 o ac \includegraphics[scale=0.7]{figures/thaitext.png}
    \item Mevalotin Pretect (40) 1x1 o pc, Mevalotin Pretect, 40, mg, 1, tablet, o, pc, 1x1, , 1x1 o pc
    \item Ridperidone (1) 0.5x1 po hs, Ridperidone, 1, mg, 0.5, tablet, po, hs, 0.5x1, , 0.5x1 po hs
    \item Cotrimazole 2 tab po od, Cotrimazole, , , 2, tablet, po, , od, , 2 tab po od
    \item Trazodone (50) 1x1 po hs, Trazodone, 50, mg, 1, tablet, po, hs, 1x1, , 1x1 po hs
    \item LoRANTA (100) 1x1, LoRANTA, 100, mg, 1, tablet, , , 1x1, , 1x1
    \item Nuelin SR (200) 1x2, Nuelin SR, 200, mg, 1, tablet, , , 1x2, , 1x2
\end{itemize}

\section{Prompts used in EX task}
\label{appendix: prompt-ex}

\paragraph{\textbf{Prompt 1}}
I will give you a table of medication data in csv format. 
Please translate and expand the information in columns named Medication ET, Unit ET, Mode ET, Instruction ET (Dose,Frequency,Duration) 

Here is the table that I want you to translate and expand: Original Text,Medication ET,Unit ET,Mode ET,"Instructions ET (Dose, Frequency, Duration)"

\begin{itemize}
    \item Amlodipine (10) 1x1 opc: Amlodipine, mg, opc, 1x1 opc
    \item ORS: ORS, , , ,
    \item paracet (500) 1 tab po prn q 4-6 hr: paracet, mg, po, 1 tab po prn q 4-6 hr
    \item Ramipril (2.5) 0.5x1 po hs: Ramipril, , po, 0.5x1 po hs
    \item VitD2 (20000) 1 tab po weekly: VitD2, , po, 1 tab po weekly
    \item Omeprazole (20) 1x1 oac: Omeprazole, mg, o, 1x1 oac
    \item Filgrastim 300 sc od start day 3-12: Filgrastim, , sc, 300 sc od start day 3-12
    \item Hidil Cap(300) 1x1: Hidil Cap, , , , 1x1
    \item Eucor Tab(20) 1x1: Eucor Tab, , , , 1x1
    \item Keflex(500) 1x4 opc: Keflex, , opc, 1x4 opc
    \item Rosuvastatin 20 1 x 1 po hs: Rosuvastatin, , po, 1x1 po hs
    \item Norfloxacin (400) 1x2 po ac: Norfloxacin, , po, 1x2 po ac
    \item Lorazepam (0.5) 1 tab po prn hs: Lorazepam, mg, po, 1 tab po prn hs
    \item Senokot 2 tabs po prn constipation hs: Senokot, , po, 2 tabs po prn constipation hs
    \item Ferli-6 1 tab po bid pc: Ferli-6, , po, 1 tab po bid pc
    \item Senokot 2 tab po hs: Senokot, , po, 2 tabs po hs
    \item Vit B co 1*3 po pc: Vit B co, , po, 1*3 po pc
    \item PROgraf Cap(1 ) 2*2 po pc: PROgraf Cap, , po, 2*2 po pc
    \item Ursolin (250) 1*3 po pc: Ursolin, , po, 1*3 po pc
    \item Paracetamol (500) 1 tab po prn q 4-6 hr: Paracetamol, mg, po, 1 tab po prn q 4-6 hr
    \item Mucotic(600) 1 \includegraphics[scale=0.7]{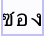} opc: Mucotic, , opc, 1 sachet opc 
    \item Clopidogrel(75) 1*1 po pc: Clopidogrel, , po, 1*1 po pc
    \item Mydocalm 1x3 po pc: Mydocalm, , po, 1x3 po pc
    \item Bisoprolol(2.5) 1/2x1 po pc: Bisoprolol, , po, 1/2x1 po pc
    \item Omeprazole(20) 1x1 po ac: Omeprazole, , po, 1x1 po ac
\end{itemize}

\paragraph{\textbf{Prompt 2}}
You are now a medication interpretator. I will give you a table of medication data in csv format. 
Please normalize the information in columns named Medication ET, Unit ET, Mode ET, Instruction ET (Dose, Frequency, Duration) based on the following instructions:

Original Text: This is the original narrative text of the medication prescription.
Medication ET: This is the medication entity type. Please interpret and put the results in a new table named "Active Ingredients EX". In case of multiple possible entries in the fields of Active Ingredients, separate the entries by ";" in the same cell. 
Unit ET: This is a unit entity type. Please interpret this and put the result in a new table named "Unit EX".
Mode ET: This is an intake route entitype type of the medication. Please interpret this and put the result in "Mode EX" column.
Instructions ET (Dose,Frequency,Duration): This is an instruction entity type. Please interpret this and put the result in "Instructions (Dose,Frequency,Duration) EX" column.

Here is the table that I want you to transform:
Original Text,Medication ET,Unit ET,Mode ET,"Instructions ET (Dose, Frequency, Duration)"

\begin{itemize}
    \item Amlodipine (10) 1x1 opc: Amlodipine, mg, opc, 1x1 opc
    \item ORS: ORS, , , ,
    \item paracet (500) 1 tab po prn q 4-6 hr: paracet, mg, po, 1 tab po prn q 4-6 hr
    \item Ramipril (2.5) 0.5x1 po hs: Ramipril, , po, 0.5x1 po hs
    \item VitD2 (20000) 1 tab po weekly: VitD2, , po, 1 tab po weekly
    \item Omeprazole (20) 1x1 oac: Omeprazole, mg, o, 1x1 oac
    \item Filgrastim 300 sc od start day 3-12: Filgrastim, , sc, 300 sc od start day 3-12
    \item Hidil Cap(300) 1x1: Hidil Cap, , , , 1x1
    \item Eucor Tab(20) 1x1: Eucor Tab, , , , 1x1
    \item Keflex(500) 1x4 opc: Keflex, , opc, 1x4 opc
    \item Rosuvastatin 20 1 x 1 po hs: Rosuvastatin, , po, 1x1 po hs
    \item Norfloxacin (400) 1x2 po ac: Norfloxacin, , po, 1x2 po ac
    \item Lorazepam (0.5) 1 tab po prn hs: Lorazepam, mg, po, 1 tab po prn hs
    \item Senokot 2 tabs po prn constipation hs: Senokot, , po, 2 tabs po prn constipation hs
    \item Ferli-6 1 tab po bid pc: Ferli-6, , po, 1 tab po bid pc
    \item Senokot 2 tab po hs: Senokot, , po, 2 tabs po hs
    \item Vit B co 1*3 po pc: Vit B co, , po, 1*3 po pc
    \item PROgraf Cap(1 ) 2*2 po pc: PROgraf Cap, , po, 2*2 po pc
    \item Ursolin (250) 1*3 po pc: Ursolin, , po, 1*3 po pc
    \item Paracetamol (500) 1 tab po prn q 4-6 hr: Paracetamol, mg, po, 1 tab po prn q 4-6 hr
    \item Mucotic(600) 1 \includegraphics[scale=0.7]{figures/thaitext2.png} opc: Mucotic, , opc, 1 sachet opc
    \item Clopidogrel(75) 1*1 po pc: Clopidogrel, , po, 1*1 po pc
    \item Mydocalm 1x3 po pc: Mydocalm, , po, 1x3 po pc
    \item Bisoprolol(2.5) 1/2x1 po pc: Bisoprolol, , po, 1/2x1 po pc
    \item Omeprazole(20) 1x1 po ac: Omeprazole, , po, 1x1 po ac
\end{itemize}

The end output should compile all results into one unified table.

\paragraph{\textbf{Prompt 3}}
You are now a medication interpretator. I will give you a table of medication data in csv format.
Please normalize the information in column named Medication ET, Unit ET, Mode ET, Instruction ET (Dose,Frequency,Duration) based on the following instructions:

Original Text: This is the original narrative text of medication prescription.
Medication ET: This is the medication entity types. Please interpret and put the results in a new table named "Active Ingredients EX". In case of multiple possible entries in the fields of Active Ingredients, separate the entries by ";" in the same cell.
Unit ET: This is a unit entity type. Please interpret this and and put the result in a new table named "Unit EX". For example, "mg" should be "milligram."
Mode ET: This is an intake route entitype type of the medication. Please interpret this and put the result in "Mode EX" column. For example, "po" should be "oral."
Instructions ET (Dose,Frequency,Duration): This is an instruction entity type. Please interpret this and put the result in "Instructions (Dose,Frequency,Duration) EX" column. For example, "1*1 po pc" should be translated into "1 tablet oral after meal once daily"

Here is the table that I want you to transform:
Original Text,Medication ET,Unit ET,Mode ET,"Instructions ET (Dose, Frequency, Duration)"

\begin{itemize}
    \item Amlodipine (10) 1x1 opc: Amlodipine, mg, opc, 1x1 opc
    \item ORS: ORS, , , ,
    \item paracet (500) 1 tab po prn q 4-6 hr: paracet, mg, po, 1 tab po prn q 4-6 hr
    \item Ramipril (2.5) 0.5x1 po hs: Ramipril, , po, 0.5x1 po hs
    \item VitD2 (20000) 1 tab po weekly: VitD2, , po, 1 tab po weekly
    \item Omeprazole (20) 1x1 oac: Omeprazole, mg, o, 1x1 oac
    \item Filgrastim 300 sc od start day 3-12: Filgrastim, , sc, 300 sc od start day 3-12
    \item Hidil Cap(300) 1x1: Hidil Cap, , , , 1x1
    \item Eucor Tab(20) 1x1: Eucor Tab, , , , 1x1
    \item Keflex(500) 1x4 opc: Keflex, , opc, 1x4 opc
    \item Rosuvastatin 20 1 x 1 po hs: Rosuvastatin, , po, 1x1 po hs
    \item Norfloxacin (400) 1x2 po ac: Norfloxacin, , po, 1x2 po ac
    \item Lorazepam (0.5) 1 tab po prn hs: Lorazepam, mg, po, 1 tab po prn hs
    \item Senokot 2 tabs po prn constipation hs: Senokot, , po, 2 tabs po prn constipation hs
    \item Ferli-6 1 tab po bid pc: Ferli-6, , po, 1 tab po bid pc
    \item Senokot 2 tab po hs: Senokot, , po, 2 tabs po hs
    \item Vit B co 1*3 po pc: Vit B co, , po, 1*3 po pc
    \item PROgraf Cap(1 ) 2*2 po pc: PROgraf Cap, , po, 2*2 po pc
    \item Ursolin (250) 1*3 po pc: Ursolin, , po, 1*3 po pc
    \item Paracetamol (500) 1 tab po prn q 4-6 hr: Paracetamol, mg, po, 1 tab po prn q 4-6 hr
    \item Mucotic(600) 1 \includegraphics[scale=0.7]{figures/thaitext2.png} opc: Mucotic, , opc, 1 sachet opc
    \item Clopidogrel(75) 1*1 po pc: Clopidogrel, , po, 1*1 po pc
    \item Mydocalm 1x3 po pc: Mydocalm, , po, 1x3 po pc
    \item Bisoprolol(2.5) 1/2x1 po pc: Bisoprolol, , po, 1/2x1 po pc
    \item Omeprazole(20) 1x1 po ac: Omeprazole, , po, 1x1 po ac
\end{itemize}

Here are some examples that you can look up to:
Original Text,Medication ET,Unit ET,Mode ET,"Instructions ET (Dose, Frequency, Duration)",Active Ingredient EX,Unit EX,Mode EX,"Instructions (Dose, Frequency, Duration) EX"

\begin{itemize}
    \item Xarator (40) 1/2x1 opc: 
    \begin{itemize}
        \item Original Text: Xarator (40) 1/2x1 opc
        \item Medication ET: Xarator
        \item Unit ET: mg
        \item Mode ET: opc
        \item Instructions ET (Dose, Frequency, Duration): 1/2x1 opc
        \item Active Ingredient EX: Simvastatin
        \item Unit EX: milligram
        \item Mode EX: Oral
        \item Instructions (Dose, Frequency, Duration) EX: 1 Tablet Oral after meal Once daily
    \end{itemize}
    
    \item Douzabox 1x2 opc: 
    \begin{itemize}
        \item Original Text: Douzabox 1x2 opc
        \item Medication ET: Douzabox
        \item Unit ET: 
        \item Mode ET: opc
        \item Instructions ET (Dose, Frequency, Duration): 1x2 opc
        \item Active Ingredient EX: Douzabox
        \item Unit EX: 
        \item Mode EX: Oral
        \item Instructions (Dose, Frequency, Duration) EX: 1 Tablet Oral after meal Twice daily
    \end{itemize}
    
    \item omeprazole (20) 1x1 po ac: 
    \begin{itemize}
        \item Original Text: omeprazole (20) 1x1 po ac
        \item Medication ET: omeprazole
        \item Unit ET: mg
        \item Mode ET: po
        \item Instructions ET (Dose, Frequency, Duration): 1x1 po ac
        \item Active Ingredient EX: Omeprazole
        \item Unit EX: milligram
        \item Mode EX: Oral
        \item Instructions (Dose, Frequency, Duration) EX: 1 Tablet Oral before meal Once daily
    \end{itemize}
    
    \item Thyrosit (50) 0.5x1 o ac \includegraphics[scale=0.7]{figures/thaitext.png}: 
    \begin{itemize}
        \item Original Text: Thyrosit (50) 0.5x1 o ac \includegraphics[scale=0.7]{figures/thaitext.png}
        \item Medication ET: Thyrosit
        \item Unit ET: mcg
        \item Mode ET: o
        \item Instructions ET (Dose, Frequency, Duration): 0.5x1 o ac \includegraphics[scale=0.7]{figures/thaitext.png}
        \item Active Ingredient EX: Thyrosit
        \item Unit EX: microgram
        \item Mode EX: Oral
        \item Instructions (Dose, Frequency, Duration) EX: 0.5 Tablet Oral before meal Once daily \includegraphics[scale=0.7]{figures/thaitext.png}
    \end{itemize}
    
    \item Mevalotin Pretect (40) 1x1 o pc: 
    \begin{itemize}
        \item Original Text: Mevalotin Pretect (40) 1x1 o pc
        \item Medication ET: Mevalotin Pretect
        \item Unit ET: mg
        \item Mode ET: o
        \item Instructions ET (Dose, Frequency, Duration): 1x1 o pc
        \item Active Ingredient EX: Pravastatin
        \item Unit EX: milligram
        \item Mode EX: Oral
        \item Instructions (Dose, Frequency, Duration) EX: 1 Tablet Oral after meal Once daily
    \end{itemize}
    
\end{itemize}

The end output should compile all results into one unified table.
\end{document}